\documentclass{article}

\PassOptionsToPackage{numbers, compress}{natbib}
\usepackage[main, final]{neurips_2026}

\usepackage[utf8]{inputenc} % allow utf-8 input
\usepackage[T1]{fontenc}    % use 8-bit T1 fonts
\usepackage[hypertexnames=false]{hyperref}       % hyperlinks
\usepackage{url}            % simple URL typesetting
\usepackage{booktabs}       % professional-quality tables
\usepackage{amsfonts}       % blackboard math symbols
\usepackage{nicefrac}       % compact symbols for 1/2, etc.
\usepackage{microtype}      % microtypography
\usepackage{xcolor}         % colors
\usepackage{colortbl}
\usepackage{graphicx}
\usepackage{subcaption}
\usepackage{multirow}
\usepackage{float}
\usepackage{makecell}
\usepackage{amsmath}
\usepackage{amssymb}
\usepackage{mathtools}
\usepackage{amsthm}
\usepackage[capitalize,noabbrev]{cleveref}
\usepackage{hyperref}

\graphicspath{{../figures/}{docs/figures/}{figs/}}

\theoremstyle{plain}
\newtheorem{theorem}{Theorem}[section]
\newtheorem{proposition}[theorem]{Proposition}
\newtheorem{lemma}[theorem]{Lemma}
\newtheorem{corollary}[theorem]{Corollary}
\theoremstyle{definition}

\newtheorem{assumption}[theorem]{Assumption}
\theoremstyle{remark}

\newcommand{\pmstd}[1]{{\scriptsize\textcolor{gray}{\(\pm\)#1}}}

\title{Accelerating Safe Reinforcement Learning with Massive Parallelism}

\author{%
  Joonyoung Lim \\
  Department of Information Convergence Engineering \\
  Pusan National University \\
  \texttt{ingyuh1015@pusan.ac.kr}
  \And
  Younghwan Yoo\thanks{Corresponding author.} \\
  School of Computer Science and Engineering \\
  Pusan National University \\
  \texttt{ymomo@pusan.ac.kr}
}
\begin{document}
\raggedbottom

\maketitle

\begin{abstract}

We investigate the problem of scaling safe reinforcement learning (safe RL) to massively parallel training regimes, where a large number of environments are executed with short rollout horizons. Although this setting improves data throughput and wall-clock efficiency, it creates a mismatch with constrained Markov decision processes (CMDPs), whose safety constraints are defined over full-episode cost returns. We address this mismatch by analyzing staggered environment resets as a phase-mixture distribution over short trajectory segments. This analysis characterizes the bias induced by phase coverage and motivates a phase-aggregated cost estimator that reconstructs episode-level cost estimates from short rollout batches. We further incorporate finite-sample uncertainty and distribution mismatch into a constraint-tightening framework, yielding a scalable Safe RL method compatible with on-policy optimization. Using an MJX-based implementation of Safety Gym, our experiments show that the proposed method achieves performance comparable to non-massively parallel implementations while substantially reducing wall-clock training time. Code is available at:~\href{https://github.com/ingyuh1015/Accelerating-Safe-RL-with-Massive-Parallelism}{\texttt{GitHub Repository}}.

\end{abstract}

\section{Introduction}\label{sec:intro}

Reinforcement learning (RL) provides a general framework for learning decision-making policies through sequential interaction with an environment and is commonly formulated as a Markov decision process (MDP). In many real-world applications, however, policies must satisfy safety requirements such as collision avoidance, energy limits, and operational constraints~\cite{10.1109/TPAMI.2024.3457538}. Safe RL addresses this challenge through constrained MDPs (CMDPs), where policy optimization is performed under explicit constraints~\cite{altman:inria-00074109}.

Recent advances in accelerator hardware have enabled massively parallel simulation, where simulation and learning can be executed with high throughput on devices such as GPUs and TPUs~\cite{koyamada2023pgx}. This regime is particularly beneficial for on-policy RL, which requires data collected from the current policy for each update and has traditionally suffered from limited sample efficiency. By collecting short rollouts concurrently from many environments, massively parallel training enables frequent policy updates with large effective batches~\cite{rudin2021learning}. It also substantially reduces wall-clock training time, making repeated runs for statistical evaluation and rapid parameter tuning feasible within a limited experimental budget~\cite{agarwal2021deep,10.5555/3618408.3618774}. Despite these advantages, Safe RL has not fully incorporated this training regime, and existing methods remain misaligned with its data-processing characteristics.

A key challenge in applying massively parallel training to Safe RL is  estimating reliable cost returns for policy update. In CMDPs, constraints are typically formulated in terms of cumulative costs, and existing Safe RL implementations commonly rely on episodic cost returns from completed or terminated episodes to balance reward maximization and constraint satisfaction~\cite{10.1109/TPAMI.2024.3457538,10.24963/ijcai.2024/913}.
However, updates in massively parallel training are highly likely to occur without any episode terminating in any of the parallel environments, particularly in tasks without early termination~\cite{bharthulwar2026staggered}. Even when an episode terminates, the cost return of the full trajectory is biased by earlier policies in two ways: \textbf{P1} the cost return includes costs incurred by actions taken under earlier policies, and \textbf{P2} these actions influence the initial-state distributions of subsequent rollouts.

In this paper, we first revisit staggered reset~\cite{bharthulwar2026staggered}, originally introduced to address the cyclical nonstationarity problem. By initializing parallel environments at different episode time steps, each update batch covers the entire episode horizon through short rollouts from different environments. This enables cost estimation by aggregating short rollout segments from different phases, each generated entirely by the
current policy, which can address \textbf{P1}. We further analyze how policy history biases the phase-initial state distribution and apply tightening to handle \textbf{P2}. The resulting framework enables existing Safe RL algorithms to train in the massively parallel regime, improving wall-clock efficiency while maintaining reliable constraint satisfaction. The main contributions are summarized as follows:

\begin{itemize}
\item We analyze staggered reset as a phase-mixture batch collection scheme, characterize how policy history biases cost returns and derive computable bounds for the return discrepancy.

\item We propose a massively parallel safe-RL framework with staggered reset that estimates cost directly from the phase-mixture batch and instantiate it with phase-initial state distribution correction estimation (PIS-DICE), a phase-initial density-ratio estimator for rollout-start state distributions.

\item We provide an implementation of Safety Gym exploration tasks using MuJoCo XLA (MJX) for end-to-end accelerator side training and evaluation, showing improved wall-clock efficiency with reliable cost control.

\end{itemize}

\section{Preliminaries}

\subsection{Constrained Markov Decision Processes}

A CMDP is defined by
\(\mathcal{M}=(\mathcal{S},\mathcal{A},P,d_0,r,c,\gamma,b_c)\),
where \(\mathcal{S}\) is the state space, \(\mathcal{A}\) is the action space,
\(P(s' \mid s,a)\) is the transition probability, \(d_0\) is the initial state
distribution, \(r:\mathcal{S}\times\mathcal{A}\to\mathbb{R}\) is the reward
function, \(c:\mathcal{S}\times\mathcal{A}\to[0,c_{\max}]\) is the cost
function, \(\gamma\in[0,1)\) is the discount factor, and \(b_c\) is the cost limit. Under a stationary policy \(\pi\), the discounted CMDP
objective is defined as follows~\cite{10.1109/TPAMI.2024.3457538,10.24963/ijcai.2024/913}:
\begin{equation}
\max_{\pi} J_r(\pi)
=
\mathbb{E}_{\pi}\!\left[\sum_{t=0}^{\infty}\gamma^t r_t\right]
\quad
\text{s.t.}
\quad
J_c(\pi)
=
\mathbb{E}_{\pi}\!\left[\sum_{t=0}^{\infty}\gamma^t c_t\right]
\le b_c .
\label{eq:discounted-cmdp-objective}
\end{equation}
\begin{assumption}[Bounded rewards, costs, and returns]
\label{assump:bounded-returns}
For \(f\in\{r,c\}\), there exists \(f_{\max}<\infty\) such that
\(|f(s,a)|\le f_{\max}\) for all
\((s,a)\in\mathcal{S}\times\mathcal{A}\).

For discounted infinite-horizon returns, the bound is \(B_f=f_{\max}/(1-\gamma)\). For finite-horizon returns, the bound is \(B_f=f_{\max}\sum_{t=0}^{H-1}\gamma^t\), with \(\gamma=1\) allowed for undiscounted episodic returns. The cost limit \(b_c\) must be expressed on the same return scale as \(J_c\).

Although we treat discounted returns, many safe RL tasks set the cost threshold as an undiscounted episodic sum. To support both, the return bound is \(B_f=f_{\max}/(1-\gamma)\) for the discounted infinite-horizon case and \(B_f=f_{\max}\sum_{t=0}^{H-1}\gamma^t\) for a finite horizon \(H\).

\end{assumption}

\subsection{Massively parallel data collection under staggered reset}
\label{subsec:staggered-reset-prelim}
In massively parallel on-policy training, each policy update uses length-\(K\) rollouts from \(E\) parallel environments. We consider episodic tasks with a fixed horizon \(H\), so the discounted returns in Eq.~\eqref{eq:discounted-cmdp-objective} truncate at \(H\). When \(K\ll H\), synchronous reset concentrates each batch on a narrow part of the task horizon, inducing a repeating distribution shift known as cyclical nonstationarity~\cite{rudin2021learning,koyamada2023pgx,bharthulwar2026staggered}. Staggered reset mitigates this by starting groups of environments at different episode steps, so that a single batch covers the whole horizon rather than a narrow slice.

On-policy RL has traditionally updated frequently from relatively small batches, whereas massively parallel simulation collects short rollouts across a large number of environments and updates from a much larger batch at far lower wall-clock cost. Safe RL fits this regime naturally, since its on-policy algorithms already rely on large batches to estimate the cost constraint with low variance. Massively parallel safe RL has been demonstrated on Isaac Gym~\cite{makoviychuk2021isaacgym}, but it has not become widespread and has not been studied from an algorithm training perspective, in part because the available tasks have short horizons which avoid the short rollout regime (\(K\ll H\)).

\subsection{Distribution correction from flow identities}
\label{subsec:dice-prelim}

Distribution correction is commonly used in off-policy and offline RL to correct the mismatch between the distribution induced by a target policy and the distribution represented by sampled transitions~\citep{nachum2019dualdice,zhang2020gendice,zhang2020gradientdice}. For the discounted infinite-horizon formulation, let \(d^\pi(s,a)\) denote the normalized discounted state--action occupancy distribution induced by a stationary policy \(\pi\), and let \(d^D(s,a)\) denote the underlying state--action sampling distribution from which the transitions in \(D\) are collected.
\begin{assumption}[Absolute continuity]
\label{assump:absolute-continuity}
\(d^{\pi}\) is absolutely continuous with respect to \(d^{D}\).
\end{assumption}

With Assumption~\ref{assump:absolute-continuity}, the density ratio is defined as
\(\rho^{\pi/D}(s,a)=\frac{d^{\pi}(s,a)}{d^{D}(s,a)}\).
Multiplying by \(\rho^{\pi/D}\) turns an expectation under \(d^{D}\) into an expectation under \(d^{\pi}\), so for a bounded function \(f\),
\begin{equation}
\mathbb{E}_{(s,a)\sim d^{\pi}}[f(s,a)]
=
\mathbb{E}_{(s,a)\sim d^{D}}\!\left[\rho^{\pi/D}(s,a)f(s,a)\right].
\label{eq:dice-change-of-measure}
\end{equation}
However, neither density is available in closed form and \(d^{D}\) is accessed only through samples, so \(\rho^{\pi/D}\) cannot be computed directly. Instead, under a stationary policy \(\pi\), the discounted state-action distribution \(d^{\pi}\) satisfies the flow identity
\begin{equation}
\begin{aligned}
d^{\pi}(s',a')
={}&(1-\gamma)d_0(s')\pi(a'\mid s')
+\gamma\,\pi(a'\mid s')\sum_{s,a}d^{\pi}(s,a)P(s'\mid s,a).
\end{aligned}
\label{eq:dice-flow-identity}
\end{equation}
The flow into \((s',a')\) splits into the initial-state contribution and the transition flow from previous state-action pairs. Substituting \(d^{\pi}=\rho^{\pi/D}d^{D}\) into Eq.~\eqref{eq:dice-flow-identity} leaves \(\rho^{\pi/D}\) as the only unknown, so distribution correction estimation (DICE) methods~\cite{nachum2019dualdice,zhang2020gendice,zhang2020gradientdice} estimate \(\rho^{\pi/D}\) by enforcing Eq.~\eqref{eq:dice-flow-identity} on samples of \(D\), without estimating the two densities separately.

\section{Methodology}\label{sec:methodology}
We formalize constraint estimation from phase-mixture batches collected from
environments with different phases. We first
analyze the return computed from the collected batch and compare it with the
corresponding current policy return. We then construct phase-wise cost
surrogates and tighten the update by considering finite-sample error and
distribution mismatch. Finally, we specialize the construction to uniform
staggered reset and introduce PIS-DICE as the distribution correction estimator
for phase-initial states.

\subsection{Phase-mixture return analysis}
\label{subsec:phase-mixture-return-analysis}
When training with parallel environments and short rollouts, early termination
of individual environments or staggered reset distributes environments over
phases \(\phi\in\{0,\ldots,H-1\}\)~\cite{agarwal2021deep,bharthulwar2026staggered}. At epoch \(n\), define the phase distribution by
\(q_n(\phi)=\frac{1}{E}|\{e:\phi_e=\phi\}|\), and let
\(E_\phi=E q_n(\phi)\) be the number of environments in phase \(\phi\).
Let \(d_{n,\phi}\) be the collected phase-initial state distribution~\cite{9992837,10.5555/3600270.3602044,9635984}. Under
policy \(\pi_n\), define
\(\zeta_{n,\phi,k}(s,a)=(P_{\pi_n}^{k} d_{n,\phi})(s)\pi_n(a|s)\). For any return function \(f \in \{r,c\}\), set
\(\ell_{f,n}(\phi,k)=\mathbb{E}_{(s,a)\sim \zeta_{n,\phi,k}}[f(s,a)]\).
The expected \(K\)-step return from rollouts initialized at phase \(\phi\) is
\begin{equation}
j_{f,n}^{K}(\phi)=\sum_{k=0}^{K-1}\gamma^k\ell_{f,n}(\phi,k).
\label{eq:phase-k-step-return}
\end{equation}

For the collected batch, the empirical \(K\)-step phase-mixture return is
\begin{equation}
\widehat{j}_{f,n}^{K}=\frac{1}{E}\sum_{e=1}^{E}\sum_{k=0}^{K-1}\gamma^k f(s_{e,\phi_e+k},a_{e,\phi_e+k}).
\label{eq:empirical-phase-mixture-return}
\end{equation}

\begin{lemma}
\label{lemma:phase-mixture-return}
For a batch with phase distribution \(q_n\), the empirical \(K\)-step return is
an unbiased estimator of the phase-mixture return, given by
\begin{equation}
\mathbb{E}\left[\widehat{j}_{f,n}^{K}\right]=\sum_{\phi=0}^{H-1}q_n(\phi)j_{f,n}^{K}(\phi).
\label{eq:phase-mixture-return}
\end{equation}
\end{lemma}

\begin{proof}
See Appendix~\ref{appendix:proof-phase-mixture-return}.
\end{proof}

The collected phase-initial distribution \(d_{n,\phi}\) may be shaped by previous
policies, which changes the state--action distribution \(\zeta_{n,\phi,k}\) and
the expected return \(\ell_{f,n}(\phi,k)\). Let
\(\bar d_{n,\phi}=P_{\pi_n}^{\phi}d_0\) denote the phase-initial distribution
induced by the current policy \(\pi_n\). The
corresponding current policy state--action distribution is
\(\bar{\zeta}_{n,\phi,k}(s,a)=\left(P_{\pi_n}^{k}\bar d_{n,\phi}\right)(s)\pi_n(a|s)\). The current policy one-step return is
\(\bar{\ell}_{f,n}(\phi,k)=\mathbb{E}_{(s,a)\sim\bar{\zeta}_{n,\phi,k}}[f(s,a)]\),
and the corresponding \(K\)-step phase-mixture return is
\begin{equation}
\bar{j}_{f,n}^{K}=\sum_{\phi=0}^{H-1}q_n(\phi)\sum_{k=0}^{K-1}\gamma^k\bar{\ell}_{f,n}(\phi,k).
\label{eq:current policy-phase-mixture-return}
\end{equation}

\begin{proposition}[Initial phase state bias bound]
\label{prop:distribution mismatch-bound}
Under Assumption~\ref{assump:bounded-returns}, for \(f\in\{r,c\}\), the
initial phase state mismatch bias can be bounded as
\begin{equation}
\left|\mathbb{E}\left[\widehat{j}_{f,n}^{K}\right]-\bar{j}_{f,n}^{K}\right|\le 2f_{\max}\sum_{\phi=0}^{H-1}q_n(\phi)\sum_{k=0}^{K-1}\gamma^k\left\|\zeta_{n,\phi,k}-\bar{\zeta}_{n,\phi,k}\right\|_{\mathrm{TV}}.
\label{eq:distribution mismatch-bound}
\end{equation}
\end{proposition}

\begin{proof}
See Appendix~\ref{appendix:proof-distribution mismatch-bound}.
\end{proof}

Proposition~\ref{prop:distribution mismatch-bound} identifies the
distributional term separating the phase-mixture return estimated from the batch
and the corresponding current policy phase-mixture return. Even in on-policy
algorithms, this correction is required when the collected phase-initial
distribution \(d_{n,\phi}\) differs from the current policy phase-initial
distribution \(\bar d_{n,\phi}\).

\subsection{Phase-wise cost surrogates and tightening}
\label{subsec:cost-surrogates-and-tightening}
Since safe RL algorithms need cost returns for dual updates or feasibility
recovery, we consider two cost surrogates from the collected phase-mixture batch. The sampled cost surrogate aggregates the
sampled cost \(c(s,a)\) with phase-coverage correction. The critic value
surrogate evaluates the cost critic \(V_{c,n}(s)\) on states sampled from the
same phase-mixture batch. Define the binary indicator \(I_{\phi,k}(t)=\mathbb{I}\{t=\phi+k\}\). The
coverage mass of timestep \(t\) and the coverage-correction weight are
\begin{equation}
m_{q_n}(t)=\sum_{\phi=0}^{H-1}q_n(\phi)\sum_{k=0}^{K-1}I_{\phi,k}(t),\qquad w_{q_n}(\phi,k)=\frac{\gamma^{\phi+k}}{m_{q_n}(\phi+k)}.
\label{eq:coverage-mass-and-weight}
\end{equation}
Assume \(m_{q_n}(t)>0\) for \(t=0,\ldots,H-1\), and let \(\Phi_n:=\{\phi:q_n(\phi)>0\}\) denote the set of observed phases. The two phase-wise surrogates are
\begin{align}
\widehat{J}_{c,n}^{K,\mathrm{sample}}
&=\sum_{\phi\in\Phi_n}q_n(\phi)\frac{1}{E_\phi}
\sum_{e:\phi_e=\phi}\sum_{k=0}^{K-1}
w_{q_n}(\phi,k)c(s_{e,\phi_e+k},a_{e,\phi_e+k}), \nonumber\\
\widehat{J}_{c,n}^{K,\mathrm{value}}
&=\sum_{\phi\in\Phi_n}q_n(\phi)\frac{1}{E_\phi K}
\sum_{e:\phi_e=\phi}\sum_{k=0}^{K-1}
V_{c,n}(s_{e,\phi_e+k}).
\label{eq:cost-surrogate-phase-wise}
\end{align}
Here \(V_{c,n}(s_t)\) predicts
\(\mathbb{E}_{\pi_n}[\sum_{\tau=t}^{\infty}\gamma^{\tau-t}c(s_\tau,a_\tau)\mid s_t]\).

\begin{proposition}[Finite-sample error bound]
\label{prop:finite-sample-bound}
Under Assumption~\ref{assump:bounded-returns}, choose either surrogate in
Eq.~\eqref{eq:cost-surrogate-phase-wise} and write it as
\(\widehat{J}_{c,n}^{K}=\frac{1}{E}\sum_{e=1}^{E}X_{e}\), where the \(X_e\) are
independent per-environment random variables. Let \(R_n\) be a deterministic range satisfying \(0\le X_e\le R_n\) for every environment \(e\), and let
\(S^2=\frac{1}{E-1}\sum_{e=1}^{E}(X_e-\widehat{J}_{c,n}^{K})^2\). By the
one-sided empirical Bernstein inequality, the finite-sample error of the cost
surrogate is bounded with probability at least
\(1-\delta\),
\begin{equation}
\mathbb{E}\left[\widehat{J}_{c,n}^{K}\right]-\widehat{J}_{c,n}^{K}
\le
\sqrt{\frac{2S^2\log(2/\delta)}{E}}
+ \frac{7R_n\log(2/\delta)}{3(E-1)}.
\label{eq:finite-sample-bound}
\end{equation}
\end{proposition}
\begin{proof}
See Appendix~\ref{appendix:proof-finite-sample-bound}.
\end{proof}

\begin{theorem}[Constraint satisfaction with tightened cost surrogate]
\label{thm:constraint-tightening}
Assume that the selected surrogate scale satisfies the distribution mismatch
upper bound
\(\displaystyle J_c(\pi_n)\le
\mathbb{E}[\widehat{J}_{c,n}^{K}]+\varepsilon_n\). Together with
Propositions~\ref{prop:distribution mismatch-bound}
and~\ref{prop:finite-sample-bound}, if the cost surrogate from the
phase-mixture batch satisfies
\begin{equation}
\widehat{J}_{c,n}^{K}+\beta_n+\varepsilon_n \le b_c,
\label{eq:tightened-surrogate-constraint}
\end{equation}
then \(J_c(\pi_n)\le b_c\) with probability at least \(1-\delta\).
\end{theorem}

\begin{proof}
See Appendix~\ref{appendix:proof-constraint-tightening}.
\end{proof}

\subsection{Uniform staggered reset}
\label{subsec:uniform-staggered-reset}
Massively parallel environment training is typically executed under JIT/XLA
compilation~\cite{koyamada2023pgx,rudin2021learning}, where keeping rollout tensors shape stable after compilation is
important. Therefore, staggered reset partitions the \(E\) environments into \(M\) phase
groups and assigns each group a fixed phase offset \(mK\) for \(m=0,\ldots,M-1\) along
the task horizon \(H=MK\). Environments that terminate early are kept in a
masked terminal state until the group reset point or the episode end point, so
no individual environment reset is performed. This group-wise distribution
yields a uniform phase allocation over non-overlapping horizon segments, with
mass \(1/M\) on each observed phase and no mass on other phases, and provides a
practical instance of the phase-mixture analysis in
Section~\ref{subsec:phase-mixture-return-analysis}.

\begin{theorem}[Uniform staggered reset induces a Ces\`aro phase mixture]
\label{thm:uniform-staggered-reset-cesaro-mixture}
Under uniform staggered reset, the phase-mixture return in
Eq.~\eqref{eq:phase-mixture-return} is the finite Ces\`aro mean of the
phase-wise returns \(j_{f,n}^{K}(\phi)\) from Eq.~\eqref{eq:phase-k-step-return},
namely \(\frac{1}{M}\sum_{m=0}^{M-1}j_{f,n}^{K}(mK)\).
\end{theorem}

\begin{proof}
See Appendix~\ref{appendix:proof-uniform-staggered-reset-cesaro-mixture}.
\end{proof}

\begin{corollary}[Ces\`aro phase-mixture return]
\label{cor:staggered-reset-cesaro-return}
Under uniform staggered reset, the phase-mixture return in
Lemma~\ref{lemma:phase-mixture-return} becomes
\begin{equation}
\mathbb{E}\left[\widehat{j}_{f,n}^{K}\right]=\frac{1}{M}\sum_{m=0}^{M-1}j_{f,n}^{K}(mK).
\label{eq:staggered-reset-cesaro-return}
\end{equation}
\end{corollary}

\begin{corollary}[Ces\`aro initial phase state bias bound]
\label{cor:staggered-reset-mismatch-bound}
Under uniform staggered reset and Assumption~\ref{assump:bounded-returns}, the
mismatch bound in Proposition~\ref{prop:distribution mismatch-bound} becomes
\begin{equation}
\left|\mathbb{E}\left[\widehat{j}_{f,n}^{K}\right]-\bar{j}_{f,n}^{K}\right|\le\frac{2f_{\max}}{M}\sum_{m=0}^{M-1}\sum_{k=0}^{K-1}\gamma^k\left\|\zeta_{n,mK,k}-\bar{\zeta}_{n,mK,k}\right\|_{\mathrm{TV}}.
\label{eq:staggered-reset-mismatch-bound}
\end{equation}
\end{corollary}

\begin{corollary}[Uniform phase-wise cost surrogate under staggered reset]
\label{cor:uniform-staggered-reset-cost-surrogate}
Under uniform staggered reset, the phase-wise surrogate in Eq.~\eqref{eq:cost-surrogate-phase-wise} becomes
\begin{align}
\widehat{J}_{c,n}^{K,\mathrm{sample}}
&=\sum_{m=0}^{M-1}\frac{1}{E_{mK}}
\sum_{e:\phi_e=mK}\sum_{k=0}^{K-1}
\gamma^{mK+k}c(s_{e,mK+k},a_{e,mK+k}), \nonumber\\
\widehat{J}_{c,n}^{K,\mathrm{value}}
&=\frac{1}{M}\sum_{m=0}^{M-1}\frac{1}{E_{mK}K}
\sum_{e:\phi_e=mK}\sum_{k=0}^{K-1}
V_{c,n}(s_{e,mK+k}).
\label{eq:staggered-reset-cost-surrogate-phasewise}
\end{align}
\end{corollary}

The corollary proofs are given in
Appendices~\ref{appendix:proof-staggered-reset-cesaro-return},
\ref{appendix:proof-staggered-reset-mismatch-bound}, and
\ref{appendix:proof-uniform-staggered-reset-cost-surrogate}, respectively.

\subsection{PIS-DICE: phase-initial state distribution correction}
\label{subsec:pis-dice}

The TV term in Eq.~\eqref{eq:staggered-reset-mismatch-bound} compares collected
state--action distributions with current policy state--action distributions at
each phase and rollout offset, which cannot be estimated directly in the
short-rollout training regime. Therefore, we extend the concept of distribution
correction from state--action distributions to phase-initial state distributions.

Under uniform staggered reset, a phase-initial pair is \(x=(s,\phi)\), where
\(s\) is the state at the start of phase \(\phi\). Let
\(\mathcal{K}_n(x^+\mid x)\) be the transition kernel obtained after executing
\(\pi_n\) for \(K\) environment steps. For consecutive phase-initial pairs used
by the estimator, \(x^+=(s^+,\phi+K)\) with \(\phi+K<H\). For every observed phase, assume \(\bar d_{n,\phi}\) is absolutely continuous
with respect to \(d_{n,\phi}\). Writing their densities with the same symbols,
define the phase-initial correction weight
\(\bar\rho_n(s,\phi)=\frac{\bar d_{n,\phi}(s)}{d_{n,\phi}(s)}\), so
\(\mathbb{E}_{s\sim d_{n,\phi}}[\bar\rho_n(s,\phi)]{=}1\). For each interior
phase \(m=0,\ldots,M-2\), the phase-initial one-block relation is
\begin{equation}
\mathbb{E}_{s\sim d_{n,mK},\,x^+\sim\mathcal{K}_n(\cdot\mid(s,mK))}
\left[
\bar\rho_n(s,mK)h(x^+)
\right]
=
\mathbb{E}_{s\sim \bar d_{n,(m+1)K}}\left[h(s,(m+1)K)\right],
\qquad \forall h\in\mathcal{H}.
\label{eq:phase-initial-flow-moment}
\end{equation}
Here \(\mathcal{H}\) is a class of bounded measurable functions. The sampled
one-block residual below is used as an empirical fitting objective for the
learned weights, while the formal bounds use the change-of-measure identity
directly. For a fixed
phase \(\phi\), let \(j_{c,n,\phi}^{K}(s)\) be the \(K\)-step expected cost return under the current policy from phase state \(s\). Averaging this
conditional expected return over the collected and current policy phase-initial state
distributions gives
\(C_{n,\phi}^{K}{=}\mathbb{E}_{s\sim d_{n,\phi}}[j_{c,n,\phi}^{K}(s)]\) and
\(\bar C_{n,\phi}^{K}{=}\mathbb{E}_{s\sim\bar d_{n,\phi}}[j_{c,n,\phi}^{K}(s)]\).
The phase-initial change of measure gives
\(\bar C_{n,\phi}^{K}-C_{n,\phi}^{K}
{=}\mathbb{E}_{s\sim d_{n,\phi}}[(\bar\rho_n(s,\phi)-1)j_{c,n,\phi}^{K}(s)]\).
Thus, we use this phase-initial correction to handle the TV term in
Eq.~\eqref{eq:staggered-reset-mismatch-bound}.

PIS-DICE uses two learned models. The ratio model outputs a nonnegative
phase-initial correction weight. The residual model maps a phase-initial pair to
a scalar whose difference across one rollout block defines an auxiliary fitting
objective for the ratio model. In the implementation, the ratio-model output is
\(\rho_n(s,\phi)=\epsilon_\rho+\operatorname{softplus}(u_\eta(s,\phi/(H-1)))\),
with \(\epsilon_\rho>0\).
Since the exact ratio \(\bar\rho_n(s,\phi)\) is not observed, the implementation
fits \(\rho_n\) using collected phase-initial samples and the states reached
after the rollout block. The residual model is used only for this empirical
fitting objective. The theoretical statements below are stated for the exact
ratio \(\bar\rho_n\), and Proposition~\ref{prop:learned-pis-dice-correction-error}
quantifies the additional error introduced by replacing \(\bar\rho_n\) with
learned weights.

For each environment \(e\), \(x_e\) denotes the phase-initial pair and \(x_e^+\)
denotes the pair obtained after the collected rollout block. Let \(f_\omega\) be
the residual-model output. The sampled residual is
\(\Delta_n(\eta,\omega)=E^{-1}\sum_{e=1}^{E}\rho_n(x_e)
(f_{\omega}(x_e)-f_{\omega}(x_e^+))\). The residual model is trained to expose
this discrepancy, and the ratio model is trained to reduce the exposed
discrepancy together with phase-wise normalization and ratio regularization. Let
\(\operatorname{sg}(\cdot)\) denote stop-gradient. The losses are
\begin{align}
\mathcal{L}_{\rho,n}(\eta)
&=
\frac{1}{E}\sum_{e=1}^{E}\rho_n(x_e)
\operatorname{sg}\!\left[f_\omega(x_e)-f_\omega(x_e^+)\right]
\nonumber\\
&\quad
+\lambda_{\mathrm{norm}}\mathcal{R}_{\mathrm{norm},n}(\rho_n)
+\lambda_{\rho2}\frac{1}{E}\sum_{e=1}^{E}(\rho_n(x_e)-1)^2 ,
\nonumber\\
\mathcal{L}_{f,n}(\omega)
&=
-\frac{1}{E}\sum_{e=1}^{E}\operatorname{sg}\!\left[\rho_n(x_e)\right]
\left(f_\omega(x_e)-f_\omega(x_e^+)\right)
+\lambda_f\frac{1}{E}\sum_{e=1}^{E}f_\omega(x_e)^2 ,
\end{align}
where \(\mathcal{R}_{\mathrm{norm},n}\) is the phase-wise normalization term,
\(\mathcal{R}_{\mathrm{norm},n}(\rho){=}\sum_{\phi\in\Phi_n}(|I_\phi|^{-1}\sum_{e\in I_\phi}\rho(x_e)-1)^2\).
After training, the raw weights
\(\rho_n(x_e)\) are clipped to \([1/\rho_{\max},\rho_{\max}]\) and normalized
within each phase. We write the final normalized weight as
\(\tilde\rho_{n,e}=\frac{|I_{\phi_e}|\rho^{\mathrm{clip}}_{n,e}}{\sum_{j\in I_{\phi_e}}\rho^{\mathrm{clip}}_{n,j}}\),
where \(\rho^{\mathrm{clip}}_{n,e}=\operatorname{clip}(\rho_n(x_e),1/\rho_{\max},\rho_{\max})\), so
\(|I_\phi|^{-1}\sum_{e\in I_\phi}\tilde\rho_{n,e}=1\) for every observed phase.
The losses above are used only to fit \(\tilde\rho_n\). The formal
bounds below use the exact change-of-measure identity for \(\bar\rho_n\) and then
account for replacing \(\bar\rho_n\) with \(\tilde\rho_n\) through an explicit
learned-ratio error term.

\begin{proposition}[PIS-DICE phase-initial discrepancy upper bound]
\label{prop:pis-dice-initial-phase-state-upper}
With the exact phase-initial ratio \(\bar\rho_n\) defined above, under
Assumption~\ref{assump:bounded-returns}, \(0\le j_{c,n,\phi}^{K}(s)\le B_c\),
the phase-wise cost discrepancy satisfies
\begin{equation}
\left|
\frac{1}{M}\sum_{m=0}^{M-1}\left(\bar C_{n,mK}^{K}-C_{n,mK}^{K}\right)
\right|
\le
\frac{B_c}{M}\sum_{m=0}^{M-1}
\mathbb{E}_{s\sim d_{n,mK}}\!\left[|\bar\rho_n(s,mK)-1|\right].
\label{eq:pis-dice-initial-phase-state-upper}
\end{equation}
\end{proposition}

\begin{proof}
See Appendix~\ref{appendix:proof-pis-dice-initial-phase-state-upper}.
\end{proof}

Proposition~\ref{prop:pis-dice-initial-phase-state-upper} gives a
phase-initial change-of-measure bound for the cost discrepancy induced by
initial phase-state mismatch. The bound replaces the state--action TV terms
over all phase offsets with a phase-initial correction term. The practical
correction therefore depends on the accuracy of the learned ratio.

\begin{proposition}[Learned PIS-DICE correction error]
\label{prop:learned-pis-dice-correction-error}
For the exact phase-initial ratio \(\bar\rho_n\), condition on a learned
nonnegative estimate \(\tilde\rho_n\). Assume the selected surrogate cost scalar \(Y_n(s,\phi)\) satisfies
\(|Y_n(s,\phi)|\le B_Y\). Then
\begin{align}
&\left|
\frac{1}{M}\sum_{m=0}^{M-1}
\mathbb{E}_{s\sim\bar d_{n,mK}}[Y_n(s,mK)]
-
\frac{1}{M}\sum_{m=0}^{M-1}
\mathbb{E}_{s\sim d_{n,mK}}[\tilde\rho_n(s,mK)Y_n(s,mK)]
\right|
\nonumber\\
&\qquad
\le
\frac{B_Y}{M}\sum_{m=0}^{M-1}
\mathbb{E}_{s\sim d_{n,mK}}
\!\left[|\bar\rho_n(s,mK)-\tilde\rho_n(s,mK)|\right].
\end{align}
\end{proposition}

\begin{proof}
See Appendix~\ref{appendix:proof-learned-pis-dice-correction-error}.
\end{proof}
\subsection{Practical update rule}
\label{subsec:practical-update-rule}

The implementation uses the uniform staggered reset form in
Eq.~\eqref{eq:staggered-reset-cost-surrogate-phasewise}. The sampled cost form
accumulates binary costs from the phase-mixture batch and compares the result
with \(b_c\) on the task cost-return scale. The critic value form evaluates the
same batch with \(V_{c,n}\) and compares it with the cost limit on the
critic-return scale.

For the selected cost surrogate, write the per-environment scalar as \(Y_{n,e}\).
For the sampled cost form,
\(Y_{n,e}{=}M\sum_{k=0}^{K-1}\gamma^{\phi_e+k}c(s_{e,\phi_e+k},a_{e,\phi_e+k})\);
for the critic value form,
\(Y_{n,e}{=}K^{-1}\sum_{k=0}^{K-1}V_{c,n}(s_{e,\phi_e+k})\). The finite-sample
term \(\beta_n\) is computed from Proposition~\ref{prop:finite-sample-bound} for
the selected scalar \(Y_{n,e}\).

PIS-DICE provides the normalized learned weight \(\tilde\rho_{n,e}\). The update
keeps the unweighted cost surrogate and uses
\(D_{n,e}{=}(\tilde\rho_{n,e}-1)Y_{n,e}\) to compute the mismatch correction
\[
\varepsilon_{n}^{\mathrm{PIS}}
=
\left[
\frac{1}{E}\sum_{e=1}^{E}D_{n,e}
+
\sqrt{\frac{2S_D^2\log(1/\delta)}{E}}
\right]_+ ,
\quad
S_D^2=\frac{1}{E-1}\sum_{e=1}^{E}\left(D_{n,e}-\frac{1}{E}\sum_{j=1}^{E}D_{n,j}\right)^2 .
\]
The policy update uses \(\varepsilon_n=\varepsilon_n^{\mathrm{PIS}}\) in Eq.~\eqref{eq:tightened-surrogate-constraint}, combining the finite-sample term \(\beta_n\) with the empirical correction computed from the learned weights. In the main experiments, PIS-DICE changes only the cost-side tightening, while the actor objective remains unchanged.

\section{Evaluation}
\label{sec:experiments}

We focus on evaluating whether the proposed massively parallel safe RL training framework can maintain reliable cost control with short, accelerator-friendly updates. The experiments first evaluate the effect of uniform staggered reset and the additional effect of tightening. We then compare the massively parallel training approach with the small-vectorized environment setting used in existing implementations in terms of both sample efficiency measured by total environment steps and wall-clock efficiency.

\paragraph{Setup}
We implement Goal tasks with Point and Car agents in MJX, following Safety Gym-style navigation task definitions~\citep{ray2019benchmarking,ji2023safetygymnasium} and the accelerator-side simulation design of JAX-based physics engines~\citep{freeman2021brax}. We further implement a locomotion task with a Laikago robot~\cite{kim2024spectral} in both MJX and MuJoCo Warp. To compare different constraint-handling paradigms, CPO, TRPO-Lag, and CRPO were chosen as representative primal-dual, Lagrangian, and primal methods, respectively~\citep{achiam2017constrained,schulman2015trpo,xu2021crpo}.
By default, the massively parallel setting uses 2,000 environments collecting a batch size of 20,000 steps per update over 500 epochs.
All experiments on the Goal tasks were conducted on Google TPU v4, while locomotion tasks were conducted on machine with NVIDIA GeForce RTX 4090 GPU.

\paragraph{Effect of staggered reset and tightening.}\label{envscale}
We first compare massively parallel training with short-rollout updates using the two initialization schemes, standard synchronous reset and staggered reset, and then evaluate the additional effect of the tightening method. Since standard initialization does not provide a finished-episode cost return at every update, the latest completed-episode cost return is reused for updates without completed episodes; we denote this estimate by LR. Staggered reset uses two cost surrogates, denoted as SC for sampled cost and CV for critic value estimates, evaluated with and without tightening. We further scale parallelism from 2k to 4k and 8k environments, so the batch size increases to 40,000 and 80,000 transitions, respectively.

Table~\ref{tab:goal-algorithm-comparison} shows the results of CPO on the Goal task across different environment scales. At 2k environments, standard initialization exceeds the cost limit, with completed episode cost returns available only sparsely. Staggered reset reduces cost returns by providing estimates from every phase-mixture batch. Without tightening, both surrogates still exceed the cost limit, even as increasing parallelism leads to higher rewards. Applying tightening further reduces costs at the expense of reward: SC keeps the reported mean episodic cost below the limit across the tested scales, whereas CV exceeds it in some configurations. The reward gap between tightened and untightened training narrows at larger scales, consistent with the smaller finite-sample bound afforded by increased sampling.

\begin{table}[!t]
\centering
\caption{Evaluation of CPO across 2k, 4k, and 8k parallel environments with a fixed rollout length of 10. Values are mean ± standard deviation of the last 10 epochs over eight seeds. \colorbox{gray!20}{Shading} indicates violation of the cost threshold (cost $>$ 25). \textbf{Bold} highlights, for each environment count, the highest reward that satisfies the cost threshold.}
\label{tab:goal-algorithm-comparison}
\small
\setlength{\tabcolsep}{4pt}
\renewcommand{\arraystretch}{1.5}
\resizebox{\textwidth}{!}{%
\begin{tabular}{ccccccccccc}
\toprule
\multirow{3}{*}{Init} & \multirow{3}{*}{Cost est.} & \multirow{3}{*}{Env} & \multicolumn{4}{c}{Point/Goal} & \multicolumn{4}{c}{Car/Goal} \\
\cmidrule(lr){4-7}\cmidrule(lr){8-11}
& & & \multicolumn{2}{c}{w/o tightening} & \multicolumn{2}{c}{w/ tightening} & \multicolumn{2}{c}{w/o tightening} & \multicolumn{2}{c}{w/ tightening} \\
\cmidrule(lr){4-5}\cmidrule(lr){6-7}\cmidrule(lr){8-9}\cmidrule(lr){10-11}
& & & Reward & Cost & Reward & Cost & Reward & Cost & Reward & Cost \\
\midrule
Standard & LR & 2K & \cellcolor{gray!20}15.13\pmstd{9.94} & \cellcolor{gray!20}55.48\pmstd{24.54} & -- & -- & \cellcolor{gray!20}29.14\pmstd{4.76} & \cellcolor{gray!20}65.87\pmstd{12.06} & -- & -- \\
\midrule
\multirow{6}{*}{Staggered}
 & \multirow{3}{*}{SC} & 2K & \cellcolor{gray!20}14.57\pmstd{2.78} & \cellcolor{gray!20}27.08\pmstd{0.84} & 7.56\pmstd{3.06} & 13.88\pmstd{1.59} & \cellcolor{gray!20}25.12\pmstd{2.11} & \cellcolor{gray!20}28.63\pmstd{1.46} & 18.52\pmstd{2.83} & 14.38\pmstd{1.18} \\
 &  & 4K & \cellcolor{gray!20}17.28\pmstd{5.37} & \cellcolor{gray!20}26.27\pmstd{1.23} & 13.42\pmstd{3.52} & 20.00\pmstd{3.49} & \cellcolor{gray!20}27.57\pmstd{1.76} & \cellcolor{gray!20}26.87\pmstd{1.18} & \textbf{24.38}\pmstd{2.04} & \textbf{18.43}\pmstd{0.75} \\
 &  & 8K & \cellcolor{gray!20}21.23\pmstd{1.05} & \cellcolor{gray!20}27.02\pmstd{1.50} & \textbf{18.83}\pmstd{0.73} & \textbf{21.19}\pmstd{2.05} & \cellcolor{gray!20}32.17\pmstd{0.95} & \cellcolor{gray!20}26.52\pmstd{1.38} & \textbf{29.57}\pmstd{0.70} & \textbf{21.06}\pmstd{0.70} \\
\cmidrule(lr){2-11}
 & \multirow{3}{*}{CV} & 2K & \cellcolor{gray!20}15.29\pmstd{3.76} & \cellcolor{gray!20}29.55\pmstd{3.43} & \textbf{13.42}\pmstd{2.06} & \textbf{24.53}\pmstd{3.93} & \cellcolor{gray!20}26.13\pmstd{2.57} & \cellcolor{gray!20}28.80\pmstd{3.16} & \textbf{22.69}\pmstd{3.00} & \textbf{22.19}\pmstd{3.23} \\
 &  & 4K & \cellcolor{gray!20}19.30\pmstd{2.27} & \cellcolor{gray!20}29.61\pmstd{3.04} & \textbf{17.40}\pmstd{2.23} & \textbf{24.57}\pmstd{2.98} & \cellcolor{gray!20}30.16\pmstd{2.04} & \cellcolor{gray!20}29.87\pmstd{3.98} & \cellcolor{gray!20}28.46\pmstd{1.37} & \cellcolor{gray!20}25.78\pmstd{3.53} \\
 &  & 8K & \cellcolor{gray!20}21.46\pmstd{1.17} & \cellcolor{gray!20}27.82\pmstd{2.25} & \cellcolor{gray!20}20.55\pmstd{2.29} & \cellcolor{gray!20}26.91\pmstd{3.21} & \cellcolor{gray!20}32.67\pmstd{0.86} & \cellcolor{gray!20}29.75\pmstd{2.06} & \cellcolor{gray!20}31.21\pmstd{1.40} & \cellcolor{gray!20}26.37\pmstd{3.17} \\
\bottomrule
\end{tabular}}
\end{table}

\begin{figure}[!t]
\centering
\begin{subfigure}{\linewidth}
\centering
\includegraphics[width=\linewidth]{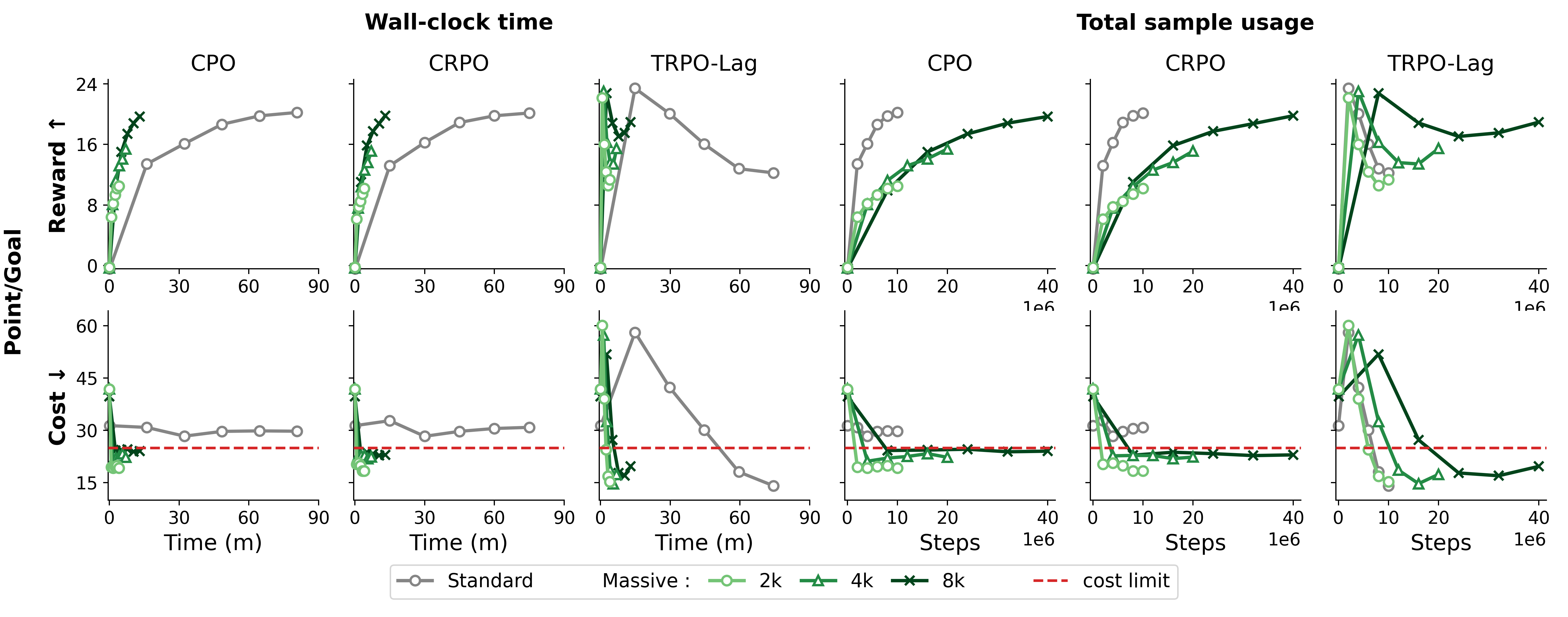}
\caption{Point/Goal. The left side shows the wall-clock wise comparison and the right side shows the total sample wise comparison.}
\label{fig:point-goal-learning}
\end{subfigure}
\vspace{0.4em}
\begin{subfigure}{\linewidth}
\centering
\includegraphics[width=\linewidth]{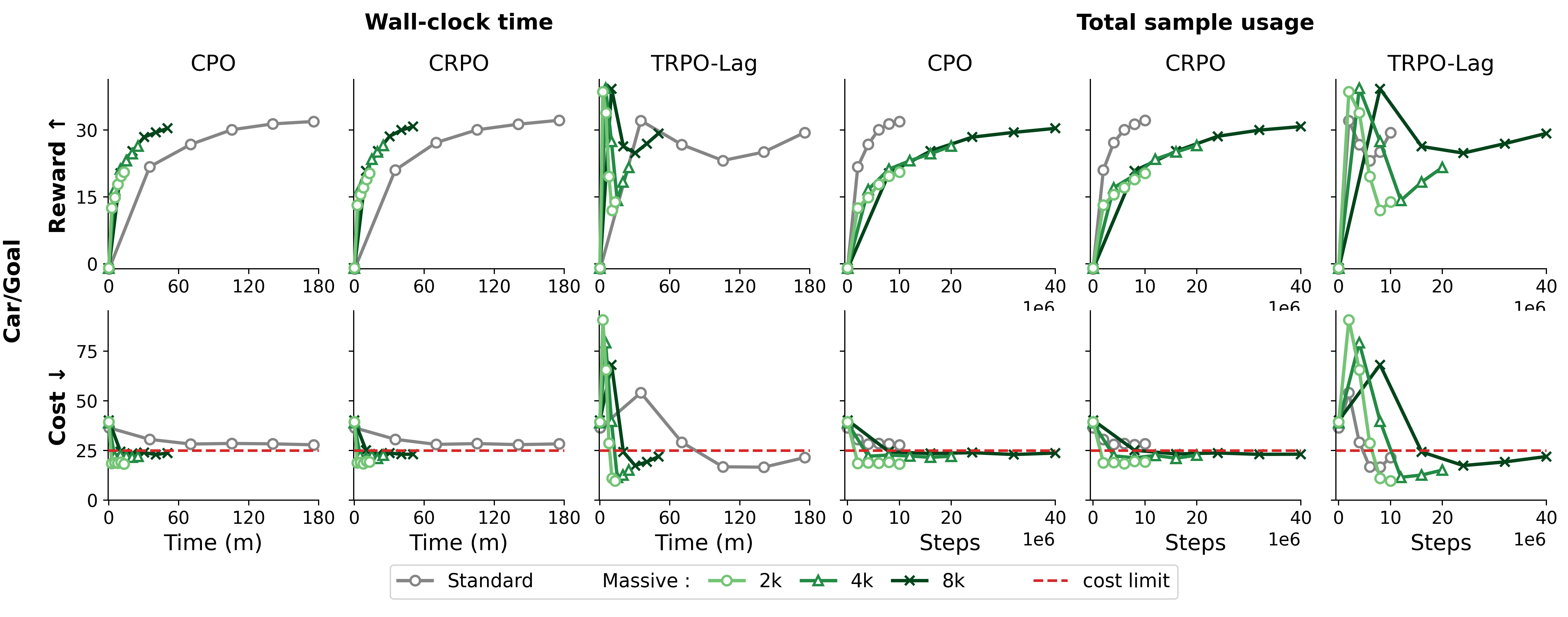}
\caption{Car/Goal. The left side shows the wall-clock wise comparison and the right side shows the total sample wise comparison.}
\label{fig:car-goal-learning}
\end{subfigure}
\caption{Wall-clock wise and total sample wise learning curves of all algorithms on the Goal task. Columns correspond to CPO, CRPO, and TRPO-Lag. Markers are shown every 100 epochs and report trailing 10-epoch averages, and the horizontal line marks the cost limit.}
\label{fig:goal-learning}
\end{figure}

\paragraph{Comparison with non-massive environment training.}
We compare the proposed massively parallel setting against the small-vectorized environment setting used in existing implementations, denoted as the standard setting. Figure~\ref{fig:goal-learning} shows the learning curves of the Goal tasks in terms of wall-clock time and total environment steps. With the same number of environment steps, the massively parallel setting generally converges to lower rewards than the standard setting using full task-horizon rollouts, which achieves better sample efficiency by providing complete task trajectories. However, the primary advantage of massively parallel training lies in wall-clock efficiency rather than sample efficiency. By processing many short rollouts from different phases concurrently on accelerators, the massively parallel setting increases transition throughput and enables more frequent policy updates within the same elapsed time. In terms of wall-clock time, training in the massively parallel setting converges to higher reward levels across algorithms as parallelism increases. In particular, 8k configuration achieves rewards comparable to the standard setting with better constraint satisfaction in substantially less training time.

\begin{figure}[!t]
\centering
\begin{subfigure}{\linewidth}
\centering
\includegraphics[width=\linewidth]{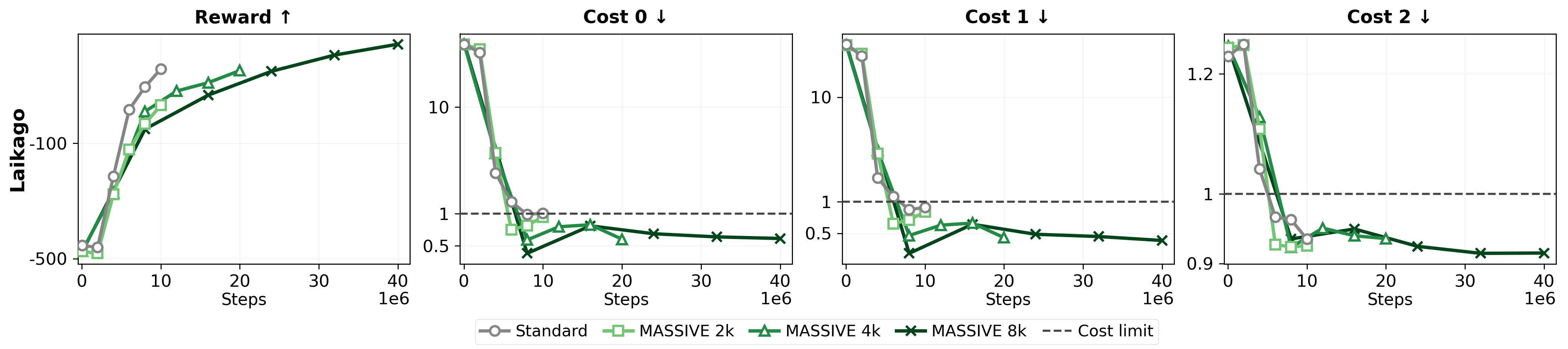}
\caption{Comparison across environment scales in terms of total environment steps, averaged over MJX and MuJoCo Warp.}
\label{fig:laikago-samples}
\end{subfigure}
\vspace{0.4em}
\begin{subfigure}{\linewidth}
\centering
\includegraphics[width=\linewidth]{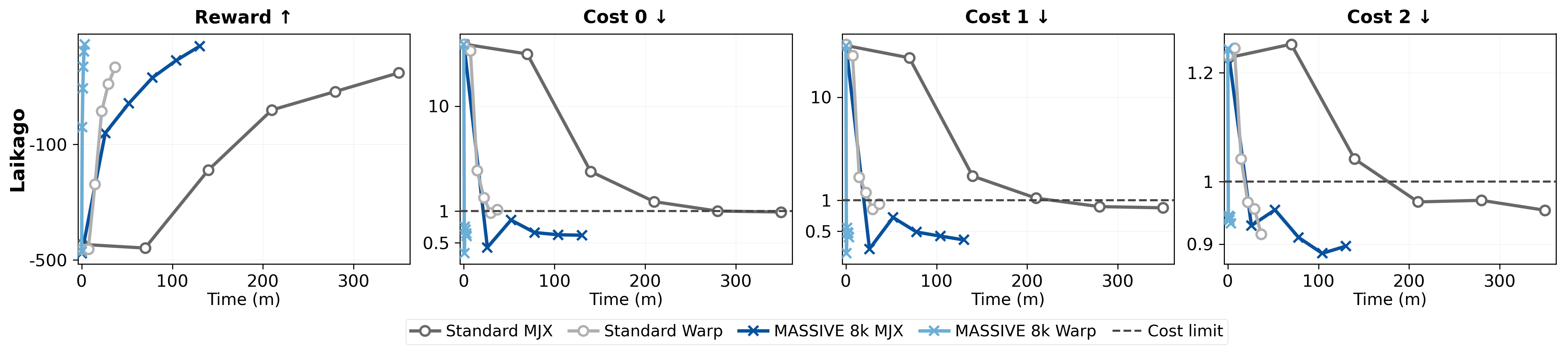}
\caption{Comparison of the standard and 8k environment settings in terms of wall-clock time using MJX and MuJoCo Warp.}
\label{fig:laikago-time}
\end{subfigure}
\caption{Learning curves of CRPO on the Laikago locomotion task. Markers are shown every 100 epochs and report trailing 10-epoch averages. Costs are normalized by their respective limits, with horizontal lines indicating the normalized cost limit of 1.}
\label{fig:laikago-learning}
\end{figure}

Figure~\ref{fig:laikago-learning} shows the learning curves of the Laikago locomotion task implemented using MJX and MuJoCo Warp. Consistent with the trends observed in the Goal tasks, reward convergence improves as parallelism increases, reaching a level comparable to the standard setting at 4K environments and surpassing it at 8K. In terms of wall-clock time, both MJX and MuJoCo Warp achieve substantially shorter training times at 8K environments than the standard setting. The difference between the two implementations further suggests that additional efficiency gains can be obtained through the choice of GPU simulation backend. Overall, these results demonstrate that massively parallel training could provide substantial wall-clock efficiency gains for safe RL research while maintaining or improving performance.

\paragraph{Ablation studies}
We conduct ablation studies to compare the effects of increasing rollout length against increasing the number of parallel environments 
and to examine how each tightening term affects training. The results are provided in Appendix~\ref{appendix:goal-ablation-studies}.

\section{Limitations}
\label{sec:limitations}
The theoretical results assume bounded costs, independent samples across environments, sufficient phase coverage, and uniform staggered resets for the simplified corollaries. Freezing terminated environments wastes some computation but preserves fixed computation shapes for efficient accelerator execution, allowing the throughput gains from massive parallelism to potentially outweigh this overhead. Nevertheless, extending the implementation to arbitrary asynchronous or highly nonuniform reset schedules remains for future work.
The practical tightening rule combines finite-sample tightening with density ratio correction, but the ablations show that most of the improvement in constraint satisfaction comes from the finite-sample term. Further evaluation on complex tasks with greater phase-initial distribution mismatch would help clarify the contribution of distribution correction. Applying tightening to multiple constraints may also become overly conservative and restrict learning, motivating further study of alternatives to tightening. The CV surrogate relies on critics trained with bootstrapped \(k\)-step targets from short rollouts and shows less consistent cost control than SC in the reported experiments. The current tightening rule does not account for critic estimation errors, and extending it to cover these errors could further increase conservatism. An alternative is to investigate distributional critics that exploit the large training batches to improve cost estimation.

\section{Conclusion}
\label{sec:conclusion}

We propose a massively parallel Safe RL framework that aligns short-rollout training with episode-level CMDP constraints through phase-covered cost estimation. By analyzing staggered reset as a phase-mixture data collection scheme, we derive cost surrogates with finite-sample and phase-initial distribution corrections, and instantiate the correction with PIS-DICE. Experiments on Safety Gym-style Goal tasks implemented in MJX and Laikago locomotion tasks implemented in both MJX and MuJoCo Warp show that the proposed framework improves wall-clock efficiency while keeping costs near their limits. These results demonstrate that safe RL can be effectively extended to massively parallel training regimes, achieving substantial gains in wall-clock efficiency while maintaining constraint satisfaction.

\newpage
\begin{ack}
This work was supported by the Institute of Information \& Communications Technology Planning \& Evaluation (IITP) grants funded by the Korea government (MSIT) through the Information Technology Research Center program (IITP-2026-RS-2023-00260098) and the Artificial Intelligence Convergence Innovation Human Resources Development program (IITP-2026-RS-2023-00254177). 

This work was also supported with Cloud TPUs from Google's TPU Research Cloud (TRC) program, and we sincerely thank the TRC team for their generous and continuous support.

\end{ack}

{\small
\bibliographystyle{plainnat}
\bibliography{ref}
}

%%%%%%%%%%%%%%%%%%%%%%%%%%%%%%%%%%%%%%%%%%%%%%%%%%%%%%%%%%%%

\newpage

\appendix

\section{Supplementary proofs and implementation details}
\label{appendix:supplementary-proofs-and-implementation}

\subsection{Proofs omitted in the main text}
\label{appendix:proofs-omitted-in-the-main-text}

\subsubsection{Proof of Lemma~\ref{lemma:phase-mixture-return}}
\label{appendix:proof-phase-mixture-return}

\begin{proof}
Condition on the phase allocation \(\{\phi_e\}_{e=1}^{E}\). For each environment \(e\) with \(\phi_e=\phi\), the distribution of \((s_{e,\phi+k},a_{e,\phi+k})\) under the rollout policy is \(\zeta_{n,\phi,k}\). Hence the definition of \(\ell_{f,n}\) gives
\begin{equation}
\mathbb{E}\left[f(s_{e,\phi+k},a_{e,\phi+k})\mid \phi_e=\phi\right]
=
\ell_{f,n}(\phi,k).
\end{equation}
Starting from the empirical phase-mixture return in Eq.~\eqref{eq:empirical-phase-mixture-return} and using linearity of expectation,
\begin{align}
\mathbb{E}\left[\widehat{j}_{f,n}^{K}\right]
&=\mathbb{E}\left[
\frac{1}{E}
\sum_{e=1}^{E}\sum_{k=0}^{K-1}
\gamma^k f(s_{e,\phi_e+k},a_{e,\phi_e+k})
\right] \\
&=\frac{1}{E}
\sum_{\phi=0}^{H-1}
\sum_{e:\phi_e=\phi}
\sum_{k=0}^{K-1}
\gamma^k
\mathbb{E}\left[f(s_{e,\phi+k},a_{e,\phi+k})\mid \phi_e=\phi\right] \\
&=\frac{1}{E}
\sum_{\phi=0}^{H-1}
\sum_{e:\phi_e=\phi}
\sum_{k=0}^{K-1}
\gamma^k\ell_{f,n}(\phi,k) \\
&=\sum_{\phi=0}^{H-1}
\frac{E_\phi}{E}
\sum_{k=0}^{K-1}
\gamma^k\ell_{f,n}(\phi,k).
\end{align}
By the definition \(q_n(\phi)=E_\phi/E\) and the phase return in Eq.~\eqref{eq:phase-k-step-return}, the final line equals
\begin{equation}
\sum_{\phi=0}^{H-1}q_n(\phi)j_{f,n}^{K}(\phi),
\end{equation}
which is Eq.~\eqref{eq:phase-mixture-return}.
\end{proof}

\subsubsection{Proof of Proposition~\ref{prop:distribution mismatch-bound}}
\label{appendix:proof-distribution mismatch-bound}

\begin{proof}
Lemma~\ref{lemma:phase-mixture-return} gives the collected phase-mixture expectation in Eq.~\eqref{eq:phase-mixture-return}. Expanding \(j_{f,n}^{K}(\phi)\) by Eq.~\eqref{eq:phase-k-step-return},
\begin{equation}
\mathbb{E}\left[\widehat{j}_{f,n}^{K}\right]
=
\sum_{\phi=0}^{H-1}q_n(\phi)
\sum_{k=0}^{K-1}\gamma^k
\mathbb{E}_{(s,a)\sim \zeta_{n,\phi,k}}[f(s,a)].
\end{equation}
The current policy phase-mixture return is Eq.~\eqref{eq:current policy-phase-mixture-return}, namely
\begin{equation}
\bar{j}_{f,n}^{K}
=
\sum_{\phi=0}^{H-1}q_n(\phi)
\sum_{k=0}^{K-1}\gamma^k
\mathbb{E}_{(s,a)\sim\bar{\zeta}_{n,\phi,k}}[f(s,a)].
\end{equation}
Subtracting the two displays and applying the triangle inequality yields
\begin{align}
\left|\mathbb{E}\left[\widehat{j}_{f,n}^{K}\right]-\bar{j}_{f,n}^{K}\right|
&=\left|
\sum_{\phi=0}^{H-1}q_n(\phi)
\sum_{k=0}^{K-1}\gamma^k
\left(
\mathbb{E}_{(s,a)\sim \zeta_{n,\phi,k}}[f(s,a)]
-
\mathbb{E}_{(s,a)\sim\bar{\zeta}_{n,\phi,k}}[f(s,a)]
\right)
\right| \\
&\le
\sum_{\phi=0}^{H-1}q_n(\phi)
\sum_{k=0}^{K-1}\gamma^k
\left|
\mathbb{E}_{(s,a)\sim \zeta_{n,\phi,k}}[f(s,a)]
-
\mathbb{E}_{(s,a)\sim\bar{\zeta}_{n,\phi,k}}[f(s,a)]
\right|.
\end{align}
For any measurable \(f\) with \(|f|\le f_{\max}\), the TV variational inequality gives
\begin{equation}
\left|\mathbb{E}_{P}[f]-\mathbb{E}_{Q}[f]\right|
=\left|\int f\,d(P-Q)\right|
\le f_{\max}\int |dP-dQ|
=2f_{\max}\|P-Q\|_{\mathrm{TV}}.
\end{equation}
Applying this inequality with \(P=\zeta_{n,\phi,k}\) and \(Q=\bar{\zeta}_{n,\phi,k}\) for every \((\phi,k)\) gives Eq.~\eqref{eq:distribution mismatch-bound}.
\end{proof}

\subsubsection{Proof of Proposition~\ref{prop:finite-sample-bound}}
\label{appendix:proof-finite-sample-bound}

\begin{proof}
After selecting one of the two cost surrogates in Eq.~\eqref{eq:cost-surrogate-phase-wise}, write it in the form assumed by the proposition,
\begin{equation}
\widehat{J}_{c,n}^{K}=\frac{1}{E}\sum_{e=1}^{E}X_e,
\qquad
0\le X_e\le R_n.
\end{equation}
The variables \(X_e\) are independent across environments under the batch construction. The one-sided empirical Bernstein inequality for independent bounded variables \citep{maurer2009empirical} states that, with probability at least \(1-\delta\),
\begin{equation}
\mathbb{E}\left[\frac{1}{E}\sum_{e=1}^{E}X_e\right]
-
\frac{1}{E}\sum_{e=1}^{E}X_e
\le
\sqrt{\frac{2S^2\log(2/\delta)}{E}}
+
\frac{7R_n\log(2/\delta)}{3(E-1)},
\end{equation}
where \(S^2=(E-1)^{-1}\sum_{e=1}^{E}(X_e-\widehat{J}_{c,n}^{K})^2\). Replacing the sample mean by \(\widehat{J}_{c,n}^{K}\) gives Eq.~\eqref{eq:finite-sample-bound}. Thus the right-hand side of Eq.~\eqref{eq:finite-sample-bound} is a valid \(\beta_n\) term for the selected surrogate.
\end{proof}

\subsubsection{Proof of Theorem~\ref{thm:constraint-tightening}}
\label{appendix:proof-constraint-tightening}

\begin{proof}
Let \(\mathcal{E}_\beta\) be the event in Proposition~\ref{prop:finite-sample-bound}. On this event, Eq.~\eqref{eq:finite-sample-bound} implies
\begin{equation}
\mathbb{E}\left[\widehat{J}_{c,n}^{K}\right]
\le
\widehat{J}_{c,n}^{K}+\beta_n.
\end{equation}
The theorem assumes the distribution mismatch upper bound
\begin{equation}
J_c(\pi_n)
\le
\mathbb{E}\left[\widehat{J}_{c,n}^{K}\right]+\varepsilon_n.
\end{equation}
Combining the two inequalities on \(\mathcal{E}_\beta\),
\begin{align}
J_c(\pi_n)
&\le
\mathbb{E}\left[\widehat{J}_{c,n}^{K}\right]+\varepsilon_n \\
&\le
\widehat{J}_{c,n}^{K}+\beta_n+\varepsilon_n.
\end{align}
If the tightened surrogate constraint in Eq.~\eqref{eq:tightened-surrogate-constraint} holds, then
\begin{equation}
J_c(\pi_n)
\le
\widehat{J}_{c,n}^{K}+\beta_n+\varepsilon_n
\le b_c.
\end{equation}
The event \(\mathcal{E}_\beta\) has probability at least \(1-\delta\), so the same probability applies to the final constraint statement.
\end{proof}

\subsubsection{Proof of Theorem~\ref{thm:uniform-staggered-reset-cesaro-mixture}}
\label{appendix:proof-uniform-staggered-reset-cesaro-mixture}

\begin{proof}
Uniform staggered reset assigns equal phase mass \(q_n(mK)=1/M\) for \(m=0,\ldots,M-1\) and zero mass to other phases. Substituting this phase distribution into the phase-mixture return from Eq.~\eqref{eq:phase-mixture-return} gives
\begin{align}
\sum_{\phi=0}^{H-1}q_n(\phi)j_{f,n}^{K}(\phi)
&=
\sum_{m=0}^{M-1}q_n(mK)j_{f,n}^{K}(mK) \\
&=
\sum_{m=0}^{M-1}\frac{1}{M}j_{f,n}^{K}(mK) \\
&=
\frac{1}{M}\sum_{m=0}^{M-1}j_{f,n}^{K}(mK).
\end{align}
The final expression is the finite Ces\`aro mean stated in
Theorem~\ref{thm:uniform-staggered-reset-cesaro-mixture}.
\end{proof}

\subsubsection{Proof of Corollary~\ref{cor:staggered-reset-cesaro-return}}
\label{appendix:proof-staggered-reset-cesaro-return}

\begin{proof}
Lemma~\ref{lemma:phase-mixture-return} gives
\begin{equation}
\mathbb{E}\left[\widehat{j}_{f,n}^{K}\right]
=
\sum_{\phi=0}^{H-1}q_n(\phi)j_{f,n}^{K}(\phi).
\end{equation}
Theorem~\ref{thm:uniform-staggered-reset-cesaro-mixture} evaluates the
right-hand side under uniform staggered reset as the Ces\`aro mean. Therefore,
\begin{equation}
\mathbb{E}\left[\widehat{j}_{f,n}^{K}\right]
=
\frac{1}{M}\sum_{m=0}^{M-1}j_{f,n}^{K}(mK),
\end{equation}
which is Eq.~\eqref{eq:staggered-reset-cesaro-return}.
\end{proof}

\subsubsection{Proof of Corollary~\ref{cor:staggered-reset-mismatch-bound}}
\label{appendix:proof-staggered-reset-mismatch-bound}

\begin{proof}
Start from the distribution mismatch bound in Eq.~\eqref{eq:distribution mismatch-bound}. Under uniform staggered reset, the only nonzero phase masses are \(q_n(mK)=1/M\) for \(m=0,\ldots,M-1\). Substituting these identities gives
\begin{align}
\left|\mathbb{E}\left[\widehat{j}_{f,n}^{K}\right]-\bar{j}_{f,n}^{K}\right|
&\le
2f_{\max}
\sum_{m=0}^{M-1}
\frac{1}{M}
\sum_{k=0}^{K-1}\gamma^k
\left\|\zeta_{n,mK,k}-\bar{\zeta}_{n,mK,k}\right\|_{\mathrm{TV}} \\
&=
\frac{2f_{\max}}{M}
\sum_{m=0}^{M-1}
\sum_{k=0}^{K-1}\gamma^k
\left\|\zeta_{n,mK,k}-\bar{\zeta}_{n,mK,k}\right\|_{\mathrm{TV}}.
\end{align}
The final expression is Eq.~\eqref{eq:staggered-reset-mismatch-bound}.
\end{proof}

\subsubsection{Proof of Corollary~\ref{cor:uniform-staggered-reset-cost-surrogate}}
\label{appendix:proof-uniform-staggered-reset-cost-surrogate}

\begin{proof}
Begin with the general phase-wise surrogates in Eq.~\eqref{eq:cost-surrogate-phase-wise}. Uniform staggered reset gives \(\phi=mK\) and \(q_n(mK)=1/M\). For every time step \(t\in\{0,\ldots,MK-1\}\), there is a unique pair \((m,k)\) with \(t=mK+k\) and \(0\le k<K\). Using the coverage mass definition in Eq.~\eqref{eq:coverage-mass-and-weight},
\begin{align}
m_{q_n}(mK+k)
&=
\sum_{\phi=0}^{H-1}q_n(\phi)
\sum_{k'=0}^{K-1}\mathbb{I}\{mK+k=\phi+k'\} \\
&=q_n(mK) \\
&=\frac{1}{M}.
\end{align}
The coverage-correction weight in Eq.~\eqref{eq:coverage-mass-and-weight} is therefore
\begin{equation}
w_{q_n}(mK,k)
=
\frac{\gamma^{mK+k}}{m_{q_n}(mK+k)}
=
M\gamma^{mK+k}.
\end{equation}
For the sampled cost surrogate in Eq.~\eqref{eq:cost-surrogate-phase-wise}, substituting \(q_n(mK)=1/M\) and the preceding weight gives
\begin{align}
\widehat{J}_{c,n}^{K,\mathrm{sample}}
&=
\sum_{m=0}^{M-1}\frac{1}{M}\frac{1}{E_{mK}}
\sum_{e:\phi_e=mK}\sum_{k=0}^{K-1}
M\gamma^{mK+k}c(s_{e,mK+k},a_{e,mK+k}) \\
&=
\sum_{m=0}^{M-1}\frac{1}{E_{mK}}
\sum_{e:\phi_e=mK}\sum_{k=0}^{K-1}
\gamma^{mK+k}c(s_{e,mK+k},a_{e,mK+k}).
\end{align}
For the critic value surrogate, Eq.~\eqref{eq:cost-surrogate-phase-wise} has no coverage-correction weight. Substituting only \(q_n(mK)=1/M\) gives
\begin{equation}
\widehat{J}_{c,n}^{K,\mathrm{value}}
=
\frac{1}{M}\sum_{m=0}^{M-1}\frac{1}{E_{mK}K}
\sum_{e:\phi_e=mK}\sum_{k=0}^{K-1}
V_{c,n}(s_{e,mK+k}).
\end{equation}
The two derived expressions are exactly Eq.~\eqref{eq:staggered-reset-cost-surrogate-phasewise}.
\end{proof}

\subsubsection{Proof of Proposition~\ref{prop:pis-dice-initial-phase-state-upper}}
\label{appendix:proof-pis-dice-initial-phase-state-upper}

\begin{proof}
For each phase \(mK\), the change-of-measure identity gives
\begin{equation}
\bar C_{n,mK}^{K}-C_{n,mK}^{K}
=
\mathbb{E}_{s\sim d_{n,mK}}
\!\left[(\bar\rho_n(s,mK)-1)j_{c,n,mK}^{K}(s)\right].
\end{equation}
Taking absolute values, applying the triangle inequality, and using
\(0\le j_{c,n,mK}^{K}(s)\le B_c\) gives
\begin{equation}
\left|\bar C_{n,mK}^{K}-C_{n,mK}^{K}\right|
\le
B_c\,
\mathbb{E}_{s\sim d_{n,mK}}
\!\left[|\bar\rho_n(s,mK)-1|\right].
\end{equation}
Applying the triangle inequality to the uniform phase average and then averaging
the preceding bound over \(m=0,\ldots,M-1\) proves
Eq.~\eqref{eq:pis-dice-initial-phase-state-upper}.
\end{proof}

\subsubsection{Proof of Proposition~\ref{prop:learned-pis-dice-correction-error}}
\label{appendix:proof-learned-pis-dice-correction-error}

\begin{proof}
For each phase \(mK\), change of measure gives
\begin{equation}
\mathbb{E}_{s\sim\bar d_{n,mK}}\!\left[Y_n(s,mK)\right]
=
\mathbb{E}_{s\sim d_{n,mK}}
\!\left[\bar\rho_n(s,mK)Y_n(s,mK)\right].
\end{equation}
Therefore, the difference between the exact-ratio correction and the
learned-ratio correction is
\begin{equation}
\mathbb{E}_{s\sim d_{n,mK}}
\!\left[(\bar\rho_n(s,mK)-\tilde\rho_n(s,mK))Y_n(s,mK)\right].
\end{equation}
Taking absolute values and using \(|Y_n(s,mK)|\le B_Y\) gives
\begin{align}
&\left|
\mathbb{E}_{s\sim\bar d_{n,mK}}\!\left[Y_n(s,mK)\right]
-
\mathbb{E}_{s\sim d_{n,mK}}
\!\left[\tilde\rho_n(s,mK)Y_n(s,mK)\right]
\right|
\nonumber\\
&\qquad\le
B_Y\,
\mathbb{E}_{s\sim d_{n,mK}}
\!\left[|\bar\rho_n(s,mK)-\tilde\rho_n(s,mK)|\right].
\end{align}
Applying the triangle inequality to the uniform phase average and then averaging
the preceding bound over \(m=0,\ldots,M-1\) gives the result.
\end{proof}

\subsection{Implementation details}
\label{appendix:implementation-details}

We implement the Safety Gym-style Goal tasks in MJX and the Laikago locomotion task in both MJX and MuJoCo Warp to support massively parallel training on accelerators.

\paragraph{Goal Tasks}
Following Safety Gym and Safety-Gymnasium~\cite{ji2023safetygymnasium,ray2019benchmarking}, we implement the Point/Goal and Car/Goal tasks with MuJoCo agent models, goals, hazards, and vases. The environments are vectorized in MJX, with rewards, costs, and observations computed at each simulation step.

\begin{table}[t]
\centering
\caption{Goal-task configurations. The cost limit specifies the maximum cumulative episodic cost.}
\label{tab:safe-rl-task-implementation}
\small
\begin{tabular}{llllll}
\toprule
Task & Agent & Horizon & Objects & Cost event & Limit \\
\midrule
Goal & Point & $1000$ & goal, $8$ hazards, $1$ vase & hazard proximity & $25$ \\
Goal & Car & $1000$ & goal, $8$ hazards, $1$ vase & hazard proximity & $25$ \\
\bottomrule
\end{tabular}
\end{table}

Both agents use two-dimensional continuous action vectors, with 12-dimensional and 24-dimensional sensor vectors for Point and Car, respectively. Observations combine these sensor vectors with three 16-bin pseudo-lidar groups for the goal, hazards, and vase, resulting in observation dimensions of 60 and 72, respectively.

Let $p_t$ and $g_t$ denote the agent and goal positions, respectively, and let $d_t=\|p_t-g_t\|_2$. The reward is defined as
\begin{equation}
r_t=d_{t-1}-d_t+\mathbb{I}\{d_t<r_{\mathrm{goal}}\},
\end{equation}
where $r_{\mathrm{goal}}$ is the goal radius. The reward consists of progress toward the goal and a unit reward upon reaching it, after which a new goal is sampled. The binary safety cost is defined as
\begin{equation}
c_t=\mathbb{I}\{\min_i\|p_t-h_i\|_2<r_{\mathrm{hazard}}\},
\end{equation}
where $h_i$ denotes the position of hazard $i$ and $r_{\mathrm{hazard}}$ is the hazard radius. Both tasks use a single safety constraint with an episodic cost limit of $25$.

\paragraph{Locomotion Task}
We adopt the Laikago locomotion task from~\cite{kim2024spectral} and implement it in both MJX and MuJoCo Warp. The robot tracks velocity commands on flat terrain. At the beginning of each episode, the forward and angular velocity commands are sampled as
\begin{equation}
v_x^{\mathrm{cmd}}\sim\mathcal{U}[-1,2],
\quad
\omega_z^{\mathrm{cmd}}\sim\mathcal{U}[-0.5,0.5],
\quad
v_y^{\mathrm{cmd}}=0.
\end{equation}
The reward penalizes squared tracking errors between commanded and actual base velocities, with an additional power regularization term.

The task uses three costs for balance, center-of-mass (CoM) height, and foot contact timing:
\begin{equation}
\begin{aligned}
c_{\mathrm{bal},t}
&=\mathbb{I}\{\theta_t>15^\circ\},\\
c_{\mathrm{hgt},t}
&=\mathbb{I}\{h_t<0.35\},\\
c_{\mathrm{foot},t}
&=\frac{1}{n_{\mathrm{legs}}}
\sum_{i=1}^{n_{\mathrm{legs}}}
\mathbb{I}\{\hat{\xi}_{t,i}\neq\xi_{t,i}\}.
\end{aligned}
\end{equation}
Here, $\theta_t$ is the angle between the robot base and world vertical axes, $h_t$ is the CoM height, and $\hat{\xi}_{t,i}$ and $\xi_{t,i}$ denote the measured and desired contact states of leg $i$, respectively. The desired contact pattern follows per-leg phase variables with a fixed frequency. The foot timing constraint uses a looser limit than the balance and height constraints, as it primarily serves as a motion regularizer.

\begin{table}[t]
\centering
\caption{Algorithm configuration. Rows with shared values span all three algorithms.}
\label{tab:goal-algorithm-config}
\footnotesize
\setlength{\tabcolsep}{4.5pt}
\renewcommand{\arraystretch}{1.05}
\begin{tabular}{lccc}
\toprule
Config & CPO & CRPO & TRPO-Lag \\
\midrule
Critic learning rate & \multicolumn{3}{c}{$10^{-3}$} \\
Discount factor & \multicolumn{3}{c}{$0.99$} \\
Reward trace parameter & \multicolumn{3}{c}{$1.0$} \\
Cost trace parameter & \multicolumn{3}{c}{$1.0$} \\
Target KL & \multicolumn{3}{c}{$0.01$} \\
CG iterations & \multicolumn{3}{c}{$15$} \\
CG damping & \multicolumn{3}{c}{$0.1$} \\
Line search coefficient & \multicolumn{3}{c}{$0.8$} \\
Rollout mode & \multicolumn{3}{c}{staggered} \\
Rollout length & \multicolumn{3}{c}{$K=10$} \\
Policy updates & \multicolumn{3}{c}{$500$} \\
Cost limit & \multicolumn{3}{c}{$25$} \\
Tightening confidence level & \multicolumn{3}{c}{$0.05$} \\
Tightening ratio clip & \multicolumn{3}{c}{$3.0$} \\
Hidden layers & \multicolumn{3}{c}{$(64,64)$} \\
Activation & \multicolumn{3}{c}{tanh} \\
Observation normalization & \multicolumn{3}{c}{true} \\
\midrule
Lagrange optimizer & -- & -- & Adam \\
Lagrange learning rate & -- & -- & $0.035$ \\
Initial Lagrange multiplier & -- & -- & $0.001$ \\
\bottomrule
\end{tabular}
\end{table}

\section{Ablation studies}
\label{appendix:goal-ablation-studies}

This appendix reports the Goal-task ablations that isolate the effects of
staggered reset, rollout length, and constraint-tightening terms. All entries
report the mean and standard deviation of the last 10 epochs over eight seeds
unless otherwise specified. SC denotes the sampled cost estimate, CV denotes the
critic value estimate.

\subsection{Rollout length}
\label{appendix:goal-fixed-n-and-horizon-ablation}

To examine the trade-off between rollout length and environment parallelism in massively parallel safe RL, Table~\ref{tab:rollout-env-batch-ablation} compares increasing batch size through longer rollouts with the environment scaling results reported in Table~\ref{tab:goal-algorithm-comparison}.

Both approaches tend to achieve higher rewards as batch size increases, although the cost limit remains violated without tightening. Once tightening is applied, all configurations satisfy the constraint, with increasing environment parallelism achieving higher rewards in most cases than extending rollout length. This difference is consistent with the larger number of samples collected per phase, which reduces the finite-sample bound and allows less conservative tightening. Moreover, the higher transition throughput obtained through greater parallelism reduces wall-clock time, making short-rollout training more effective in terms of both reward performance and training efficiency.

% Preamble: \usepackage{makecell}

\begin{table}[H]
\centering
\caption{Comparison of increasing rollout length and environment scale at the same batch size using CPO with the SC surrogate. Values are mean $\pm$ standard deviation of the last 10 updates over eight seeds. \colorbox{gray!20}{Shading} indicates cost $>$ 25. \textbf{Bold} highlights the highest feasible reward and its corresponding cost and time for each task and batch size.}
\label{tab:rollout-env-batch-ablation}
\small
\setlength{\tabcolsep}{4pt}
\renewcommand{\arraystretch}{1.5}
\resizebox{\textwidth}{!}{%
\begin{tabular}{ccccccccc|ccccccc}
\toprule
\multirow{3}{*}{Task} & \multirow{3}{*}{Batch size} & \multicolumn{7}{c|}{Fixed envs (2k)} & \multicolumn{7}{c}{Fixed rollout (10)} \\
\cmidrule(lr){3-9}\cmidrule(lr){10-16}
 & & \multirow{2}{*}{Rollout} & \multicolumn{3}{c}{w/o tightening} & \multicolumn{3}{c|}{w/ tightening} & \multirow{2}{*}{Envs} & \multicolumn{3}{c}{w/o tightening} & \multicolumn{3}{c}{w/ tightening} \\
\cmidrule(lr){4-6}\cmidrule(lr){7-9}\cmidrule(lr){11-13}\cmidrule(lr){14-16}
 & & & Reward & Cost & Time & Reward & Cost & Time & & Reward & Cost & Time & Reward & Cost & Time \\
\midrule
\multirow{3}{*}{\rotatebox[origin=c]{90}{Point/Goal}}
 & 20k & 10 & \cellcolor{gray!20}\shortstack{14.57\\\pmstd{2.78}} & \cellcolor{gray!20}\shortstack{27.08\\\pmstd{0.84}} & \cellcolor{gray!20}4.10 & \shortstack{\textbf{7.56}\\\pmstd{3.06}} & \shortstack{\textbf{13.88}\\\pmstd{1.59}} & \textbf{4.09} & 2k & -- & -- & -- & -- & -- & -- \\
 & 40k & 20 & \cellcolor{gray!20}\shortstack{17.96\\\pmstd{1.12}} & \cellcolor{gray!20}\shortstack{27.05\\\pmstd{1.10}} & \cellcolor{gray!20}7.77 & \shortstack{\textbf{13.96}\\\pmstd{1.41}} & \shortstack{\textbf{15.05}\\\pmstd{0.63}} & \textbf{7.88} & 4k & \cellcolor{gray!20}\shortstack{17.28\\\pmstd{5.37}} & \cellcolor{gray!20}\shortstack{26.27\\\pmstd{1.23}} & \cellcolor{gray!20}6.84 & \shortstack{13.42\\\pmstd{3.52}} & \shortstack{20.00\\\pmstd{3.49}} & 7.09 \\
 & 80k & 40 & \cellcolor{gray!20}\shortstack{20.46\\\pmstd{0.48}} & \cellcolor{gray!20}\shortstack{27.18\\\pmstd{0.91}} & \cellcolor{gray!20}15.91 & \shortstack{16.43\\\pmstd{0.81}} & \shortstack{16.36\\\pmstd{0.39}} & 16.20 & 8k & \cellcolor{gray!20}\shortstack{21.23\\\pmstd{1.05}} & \cellcolor{gray!20}\shortstack{27.02\\\pmstd{1.50}} & \cellcolor{gray!20}12.87 & \shortstack{\textbf{18.83}\\\pmstd{0.73}} & \shortstack{\textbf{21.19}\\\pmstd{2.05}} & \textbf{13.03} \\
\midrule
\multirow{3}{*}{\rotatebox[origin=c]{90}{Car/Goal}}
 & 20k & 10 & \cellcolor{gray!20}\shortstack{25.12\\\pmstd{2.11}} & \cellcolor{gray!20}\shortstack{28.63\\\pmstd{1.46}} & \cellcolor{gray!20}12.92 & \shortstack{\textbf{18.52}\\\pmstd{2.83}} & \shortstack{\textbf{14.38}\\\pmstd{1.18}} & \textbf{13.03} & 2k & -- & -- & -- & -- & -- & -- \\
 & 40k & 20 & \cellcolor{gray!20}\shortstack{29.11\\\pmstd{1.69}} & \cellcolor{gray!20}\shortstack{26.45\\\pmstd{0.77}} & \cellcolor{gray!20}25.64 & \shortstack{22.29\\\pmstd{2.90}} & \shortstack{15.49\\\pmstd{0.66}} & 25.75 & 4k & \cellcolor{gray!20}\shortstack{27.57\\\pmstd{1.76}} & \cellcolor{gray!20}\shortstack{26.87\\\pmstd{1.18}} & \cellcolor{gray!20}24.82 & \shortstack{\textbf{24.38}\\\pmstd{2.04}} & \shortstack{\textbf{18.43}\\\pmstd{0.75}} & \textbf{24.96} \\
 & 80k & 40 & \cellcolor{gray!20}\shortstack{32.55\\\pmstd{0.52}} & \cellcolor{gray!20}\shortstack{26.54\\\pmstd{0.74}} & \cellcolor{gray!20}51.22 & \shortstack{26.76\\\pmstd{1.16}} & \shortstack{16.69\\\pmstd{0.70}} & 51.32 & 8k & \cellcolor{gray!20}\shortstack{32.17\\\pmstd{0.95}} & \cellcolor{gray!20}\shortstack{26.52\\\pmstd{1.38}} & \cellcolor{gray!20}50.24 & \shortstack{\textbf{29.57}\\\pmstd{0.70}} & \shortstack{\textbf{21.06}\\\pmstd{0.70}} & \textbf{50.53} \\
\bottomrule
\end{tabular}%
}
\end{table}

\subsection{Tightening term ablation}
\label{appendix:goal-epsilon-only-ablation}

To examine the contribution of each tightening term to constraint satisfaction, Table~\ref{tab:goal-ablation-epsilon-only} compares CPO with the finite-sample term $\beta$ and the phase-initial distribution correction term $\varepsilon$, applied individually and jointly to the SC and CV surrogates across environment scales from 2k to 32k.

The results show that applying the finite-sample term $\beta$ accounts for most of the improvement in constraint satisfaction, whereas $\varepsilon$ alone has little effect on cost returns. For the SC surrogate, $\beta$ satisfies the constraint across all environment scales, and combining both terms provides further cost reductions in some configurations. However, the CV surrogate continues to violate the constraint in several configurations even with both terms, since critic approximation errors are not covered by the finite-sample bound. The limited effect of $\varepsilon$ in these experiments also indicates that further evaluation on tasks with greater phase-initial distribution mismatch is needed to assess the contribution of distribution correction.

\begin{table}[H]
\centering
\caption{Ablation of finite-sample tightening ($\beta$) and phase-initial distribution correction ($\varepsilon$) across environment scales using SC and CV surrogates. Values are mean $\pm$ standard deviation of the last 10 epochs over eight seeds. \colorbox{gray!20}{Shading} indicates cost $>$ 25.}
\label{tab:goal-ablation-epsilon-only}
\small
\setlength{\tabcolsep}{4pt}
\renewcommand{\arraystretch}{1.5}
\resizebox{\textwidth}{!}{%
\begin{tabular}{lrcccccccc}
\toprule
\multirow{2}{*}{Cost est.} & \multirow{2}{*}{Envs} & \multicolumn{2}{c}{No tightening} & \multicolumn{2}{c}{$\varepsilon$ only} & \multicolumn{2}{c}{$\beta$ only} & \multicolumn{2}{c}{Both ($\varepsilon+\beta$)} \\
\cmidrule(lr){3-4}\cmidrule(lr){5-6}\cmidrule(lr){7-8}\cmidrule(lr){9-10}
 & & Reward & Cost & Reward & Cost & Reward & Cost & Reward & Cost \\
\midrule
\multirow{5}{*}{SC}
 & 2k & \cellcolor{gray!20}14.57 \pmstd{2.78} & \cellcolor{gray!20}27.08 \pmstd{0.84} & \cellcolor{gray!20}12.87 \pmstd{3.83} & \cellcolor{gray!20}25.77 \pmstd{1.87} & 9.55 \pmstd{4.08} & 17.43 \pmstd{2.39} & 7.56 \pmstd{3.06} & 13.88 \pmstd{1.59} \\
 & 4k & \cellcolor{gray!20}17.28 \pmstd{5.37} & \cellcolor{gray!20}26.27 \pmstd{1.23} & \cellcolor{gray!20}16.44 \pmstd{3.21} & \cellcolor{gray!20}27.10 \pmstd{2.27} & 12.71 \pmstd{1.99} & 18.19 \pmstd{1.91} & 13.42 \pmstd{3.52} & 20.00 \pmstd{3.49} \\
 & 8k & \cellcolor{gray!20}21.23 \pmstd{1.05} & \cellcolor{gray!20}27.02 \pmstd{1.50} & \cellcolor{gray!20}21.27 \pmstd{0.69} & \cellcolor{gray!20}27.02 \pmstd{1.62} & 18.68 \pmstd{1.43} & 22.69 \pmstd{1.70} & 18.83 \pmstd{0.73} & 21.19 \pmstd{2.05} \\
 & 16k & \cellcolor{gray!20}23.07 \pmstd{0.46} & \cellcolor{gray!20}25.73 \pmstd{0.84} & \cellcolor{gray!20}23.09 \pmstd{0.48} & \cellcolor{gray!20}25.72 \pmstd{1.20} & 22.09 \pmstd{0.55} & 21.86 \pmstd{1.02} & 22.30 \pmstd{0.68} & 22.66 \pmstd{1.41} \\
 & 32k & \cellcolor{gray!20}23.81 \pmstd{0.64} & \cellcolor{gray!20}25.07 \pmstd{0.33} & \cellcolor{gray!20}24.37 \pmstd{0.54} & \cellcolor{gray!20}25.88 \pmstd{0.74} & 23.48 \pmstd{0.70} & 23.19 \pmstd{0.42} & 23.43 \pmstd{0.70} & 23.06 \pmstd{0.59} \\
\midrule
\multirow{5}{*}{CV}
 & 2k & \cellcolor{gray!20}15.29 \pmstd{3.76} & \cellcolor{gray!20}29.55 \pmstd{3.43} & \cellcolor{gray!20}16.41 \pmstd{2.95} & \cellcolor{gray!20}26.62 \pmstd{1.28} & 13.49 \pmstd{2.16} & 24.60 \pmstd{5.18} & 13.42 \pmstd{2.06} & 24.53 \pmstd{3.93} \\
 & 4k & \cellcolor{gray!20}19.30 \pmstd{2.27} & \cellcolor{gray!20}29.61 \pmstd{3.04} & \cellcolor{gray!20}19.91 \pmstd{1.09} & \cellcolor{gray!20}28.82 \pmstd{2.96} & 17.10 \pmstd{1.24} & 23.31 \pmstd{1.66} & 17.40 \pmstd{2.23} & 24.57 \pmstd{2.98} \\
 & 8k & \cellcolor{gray!20}21.46 \pmstd{1.17} & \cellcolor{gray!20}27.82 \pmstd{2.25} & \cellcolor{gray!20}21.71 \pmstd{1.58} & \cellcolor{gray!20}27.16 \pmstd{2.66} & \cellcolor{gray!20}20.43 \pmstd{2.75} & \cellcolor{gray!20}26.16 \pmstd{2.34} & \cellcolor{gray!20}20.55 \pmstd{2.29} & \cellcolor{gray!20}26.91 \pmstd{3.21} \\
 & 16k & \cellcolor{gray!20}23.10 \pmstd{0.87} & \cellcolor{gray!20}26.78 \pmstd{2.76} & \cellcolor{gray!20}23.38 \pmstd{0.59} & \cellcolor{gray!20}26.07 \pmstd{1.56} & \cellcolor{gray!20}23.05 \pmstd{0.81} & \cellcolor{gray!20}25.55 \pmstd{1.69} & 22.79 \pmstd{1.18} & 23.94 \pmstd{2.26} \\
 & 32k & \cellcolor{gray!20}24.82 \pmstd{0.36} & \cellcolor{gray!20}26.82 \pmstd{1.42} & \cellcolor{gray!20}24.39 \pmstd{0.73} & \cellcolor{gray!20}25.37 \pmstd{2.50} & \cellcolor{gray!20}24.42 \pmstd{0.47} & \cellcolor{gray!20}25.28 \pmstd{1.84} & 24.41 \pmstd{0.86} & 24.92 \pmstd{1.61} \\
\bottomrule
\end{tabular}%
}
\end{table}

%%%%%%%%%%%%%%%%%%%%%%%%%%%%%%%%%%%%%%%%%%%%%%%%%%%%%%%%%%%%

\newpage

\end{document}